\documentclass[lettersize,journal]{IEEEtran}
\IEEEoverridecommandlockouts
\usepackage{amsmath,amssymb,amsfonts}
\usepackage{algorithmic}
\usepackage{array}
\usepackage{graphicx}
\usepackage{textcomp}
\usepackage{stfloats}
\usepackage{url}
\usepackage{verbatim}
\usepackage[caption=false,font=normalsize,labelfont=sf,textfont=sf]{subfig}

\usepackage[utf8]{inputenc}
\usepackage[colorlinks=true, linkcolor=red, citecolor=green, urlcolor=magenta]{hyperref}
\usepackage{xcolor}
\usepackage[numbers, sort&compress]{natbib}
\usepackage{fontawesome}
\usepackage{pifont}
\usepackage{algorithm}
\usepackage{multirow}
\usepackage{booktabs}
\def\BibTeX{{\rm B\kern-.05em{\sc i\kern-.025em b}\kern-.08em
    T\kern-.1667em\lower.7ex\hbox{E}\kern-.125emX}}

\begin{document}
\title{Boosting Active Defense Persistence: A Two-Stage Defense Framework Combining Interruption and Poisoning Against Deepfake \\
}

\author{Hongrui Zheng$^{1}$,
        Yuezun Li$^{2}$,
        Liejun Wang$^{1,3,4}$ \textsuperscript{\faEnvelopeO},
        Yunfeng Diao$^{5}$,
        Zhiqing Guo$^{1,3,4}$ \textsuperscript{\faEnvelopeO}\\
    \textsuperscript{1}School of Computer Science and Technology, Xinjiang University, Urumqi, China \\
    \textsuperscript{2}Department of Computer Science and Technology, Ocean University of China, Qingdao, China \\
    \textsuperscript{3}Xinjiang Multimodal Intelligent Processing and Information Security \\
    Engineering Technology Research Center, Urumqi, China\\
    \textsuperscript{4}Silk Road Multilingual Cognitive Computing lnternational Cooperation Joint Laboratory,\\
    Xinjiang University, Urumqi, China \\
    
    \textsuperscript{5}School of Computer Science and Information Engineering, Hefei University of Technology, Hefei, China\\

    \text{107552301310@stu.xju.edu.cn}, \text{liyuezun@ouc.edu.cn},
    \text{wljxju@xju.edu.cn},\\
\text{diaoyunfeng@hfut.edu.cn}, \text{guozhiqing@xju.edu.cn}

\thanks{\textsuperscript{\faEnvelopeO} Corresponding authors.}
\thanks{This work was supported in part by the National Natural Science Foundation of China under Grant 62462060, Grant 62302427, and Grant 62472368, in part by the Natural Science Foundation of Xinjiang Uygur Autonomous Region under Grant 2023D01C175, in part by the Central Government Guides Local Science and Technology Development Fund Projects under Grant ZYYD2026ZY21.}}

\markboth{IEEE TRANSACTIONS ON INFORMATION FORENSICS AND SECURITY}%
{How to Use the IEEEtran \LaTeX \ Templates}
\maketitle

\begin{abstract}
    Active defense strategies have been developed to counter the threat of deepfake technology. However, a primary challenge is their lack of persistence, as their effectiveness is often short-lived. Attackers can bypass these defenses by simply collecting protected samples and retraining their models. This means that static defenses inevitably fail when attackers retrain their models, which severely limits practical use.
    We argue that an effective defense not only distorts forged content but also blocks the model's ability to adapt, which occurs when attackers retrain their models on protected images. To achieve this, we propose an innovative Two-Stage Defense Framework (TSDF). Benefiting from the intensity separation mechanism designed in this paper, the framework uses dual-function adversarial perturbations to perform two roles. First, it can directly distort the forged results. Second, it acts as a poisoning vehicle that disrupts the data preparation process essential for an attacker's retraining pipeline. By poisoning the data source, TSDF aims to prevent the attacker's model from adapting to the defensive perturbations, thus ensuring the defense remains effective long-term.
    Comprehensive experiments show that the performance of traditional interruption methods degrades sharply when these methods are subjected to adversarial retraining. However, our framework shows a strong dual defense capability, which can improve the persistence of active defense. Our code will be available at https://github.com/vpsg-research/TSDF.
\end{abstract}
\section{Introduction}

\begin{figure}[htbp]
    \centering
    \includegraphics[scale=0.19]{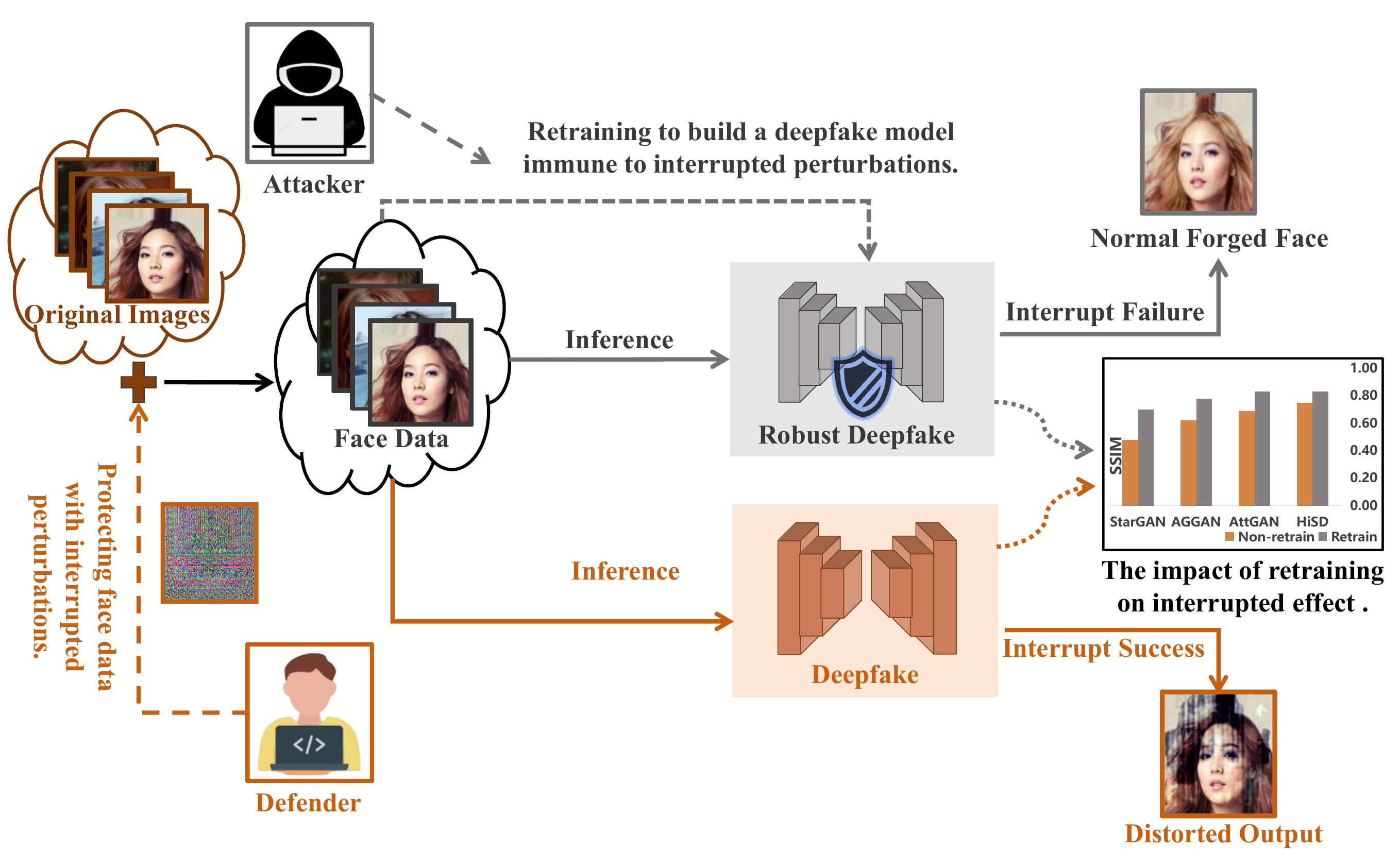}  
    \caption{Illustration of an interruption-based defense and its failure to retrain. 
    The lower path demonstrates an effective interruption process, where protected face data leads to a distorted output from the deepfake model.
    The upper path illustrates a critical flaw in this interruption-only defense. Attackers can bypass the defense by retraining their model on the protected images. This adaptation makes the model immune to the interruption, ultimately causing the defense to fail.}
    \label{fig:01}
\end{figure}

\IEEEPARstart{I}{n} recent years, deepfake technology, primarily driven by Generative Adversarial Networks (GANs), has made significant advancements in generating highly realistic and convincing forged images and videos \cite{b1}. The high fidelity of this generated media usually makes it extremely difficult for human eyes to distinguish it from the real content, and the increasing availability of these tools increases the possibility of their abuse. With the proliferation of malicious applications such as political disinformation, identity theft and financial fraud, these behaviors are seriously infringing on personal privacy, undermining social stability and eroding the credibility of public institutions \cite{b2}.

\begin{figure*}[htbp]
    \centering
    \includegraphics[scale=0.245]{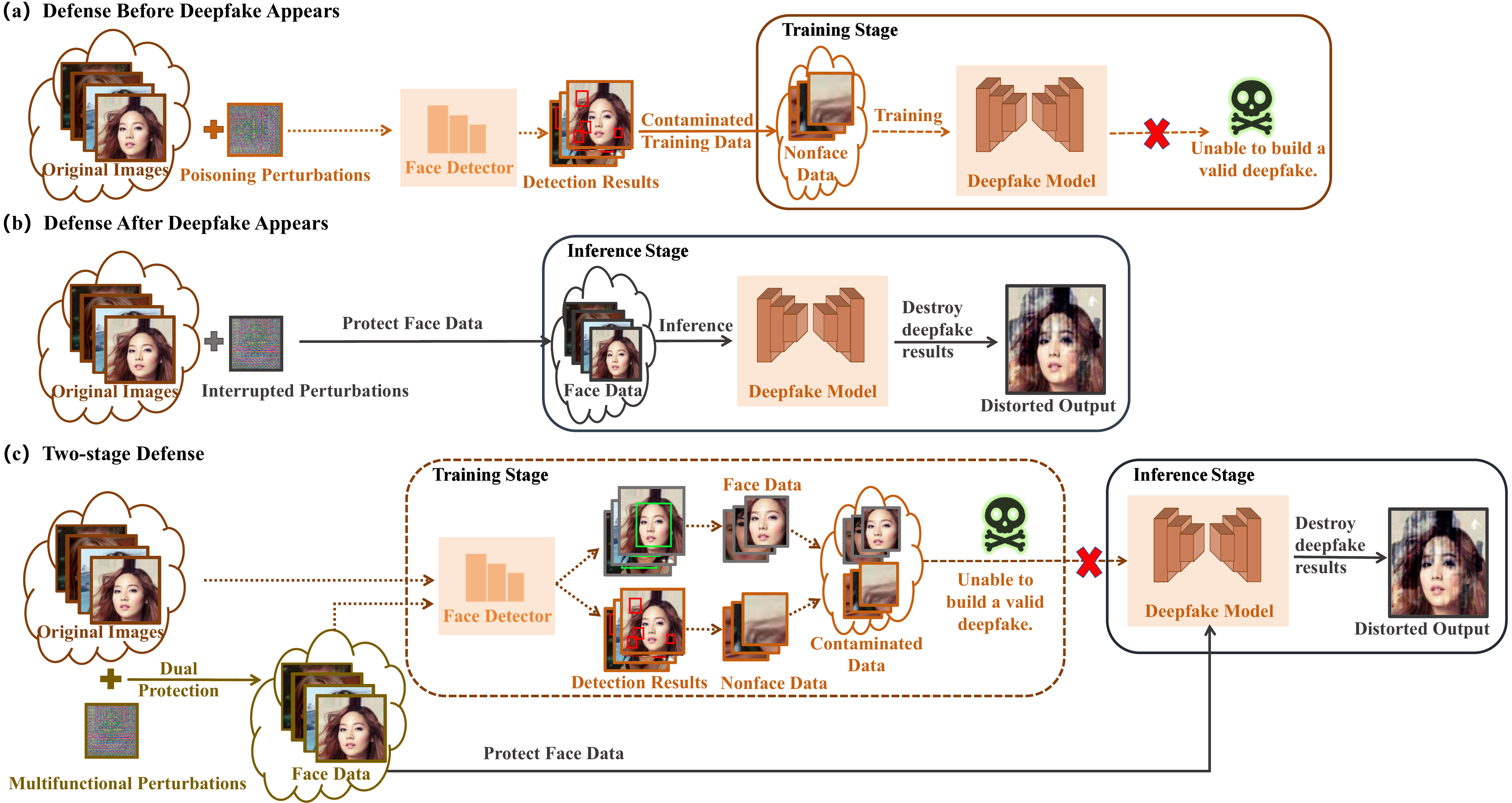}  
    \caption{Comparison of three active defense strategies. (a) Defense Before Deepfake Appears: this path shows a poisoning strategy applied before deepfake model training. The defender poisons the original training data by interfering with face detection. This prevents the attacker's model from learning effective facial features, causing the model creation to fail. (b) Defense After Deepfake Appears: this path shows an interruption strategy applied at model inference stage. The defender adds interruption perturbations to the face images that need to be protected. When these protected images are fed into a trained deepfake model, a distorted output is produced. (c) Two-Stage Defense: this path shows the framework proposed in this paper, which combines the previous two strategies in one perturbation. The framework applies dual protection to the original images. This allows it to both distort the forged result at the inference stage and poison the dataset in the training stage, thus improving the persistence of active defense.}
    \label{fig:7}
\end{figure*}

To counter these threats, researchers have proposed two types of defense strategies: passive detection and active defense. Passive detection aims to identify a forged image after its creation by analyzing a wide array of forged traces, such as GAN fingerprints, compression artifacts, or inconsistent biological signals \cite{b3, b4, b5, b6, b10, b42, b43}. However, this method cannot prevent the production of forgery media in advance. Considering the characteristics of social media, forgeries can spread quickly and cause irreversible damage before they are detected and marked \cite{b7, b8}. Furthermore, passive detection methods create a continuous arms race, as new generative models are often explicitly trained to eliminate the artifacts that detectors search for, requiring constant updates to detection techniques. This defect of passive response highlights the necessity of adopting a more active strategy. Consequently, active defense has gained increasing attention, as it aims to prevent malicious content from being created or spread in the first place \cite{b11, b12, b13, b40, b41}. The core idea is to intervene before or during content generation to disrupt the forgery process itself. Common methods fall into two main categories: interruption and data poisoning. For the former, the goal of interruption is to distort the forged images at the moment of creation \cite{b14, b39}. For example, CMUA \cite{b15} embeds perturbations into images to make the final output visibly distorted or unusable. TAFIM method \cite{b9} prevents the face from being tampered by changing the model output into a pure color image, which can also be used to identify manipulation techniques. For the latter, the goal of data poisoning is to disrupt the deepfake model in the training stage. Some methods, such as Landmark Breaker \cite{b19} and Face Poison \cite{b22}, achieve this by introducing contaminated data to disrupt key pre-processing steps like facial alignment, thus degrading the quality of the trained model \cite{b16}.

While the interruption strategies are effective in the inference stage, they share the critical challenge of lacking persistence. As illustrated in the upper path of Fig. \ref{fig:01}, this weakness becomes apparent when an attacker collects a set of protected images and incorporates them into their training dataset. For example, a comparison of the Structural Similarity Index Measure(SSIM) shows that the output from a retrained model is much less distorted than the output from the original model. The core of the problem lies in the pre-processing stage. Conventional interruption perturbations do not interfere with face detectors, meaning that an attacker's model still receives the protected face as a valid training input. Consequently, from the model's perspective, these perturbations are not a disruption to be ignored, but simply another data feature to be learned. Given that the model's objective is to accurately reproduce its training data, it naturally learns to replicate the perturbation patterns as part of the face itself. This adaptation allows the model to become immune to the perturbations, which significantly reduces the defense's effectiveness and exposes a fundamental flaw in existing static defense strategies \cite{b17, b24}. This shows that the defense is effective only upon initial deployment and fails to provide persistent protection against an adaptive attacker.

To overcome this cycle of defense adaptation and build a new paradigm for achieving a more persistent defense, we argue that the defense system should have dual capabilities. That is, it not only needs to interrupt the current forgery content, but also needs to destroy the attacker's subsequent adaptability. To this end, we propose a Two-Stage Defense Framework. As shown in Fig. \ref{fig:7}, the framework's key insight is the synergistic combination of two strategies in a single mechanism. This combination is efficiently managed by our proposed intensity separation mechanism, which ensures the two defensive functions can operate cooperatively without conflict. In this design, the poisoning mechanism acts as a persistent shield for the interruption function. In the inference stage, the perturbations act as an interruption tool to distort the output \cite{b15}. In the training stage, it also serves as a poisoning perturbation to interfere with face localization \cite{b22}. This contaminates the dataset used by the attacker for retraining, thus protecting the interruption function from being bypassed.

The main contributions of this paper are summarized as follows:
\begin{itemize}

    \item We identify and analyze the critical persistence vulnerability in existing active defense methods.
    \item We design an innovative Two-Stage Defense Framework that effectively addresses this challenge by combining interruption and poisoning.
    \item We propose an intensity separation mechanism that allows both defensive functions to work together efficiently.
    
\end{itemize}

\section{RELATED WORKS}

\subsection{Single-Function Perturbation Methods}\label{AA}

Data poisoning is an active defense strategy that interferes with a deepfake model's learning process by injecting perturbations into the training data. Prior works have explored various approaches that primarily target specific pre-processing steps to contaminate the training data pipeline. Some methods, such as Landmark Breaker \cite{b19}, focus on a upstream task by disrupting the extraction of facial key points, which aims to degrade the final model's quality by providing it with poorly aligned facial data for training. However, this approach targets a upstream task that is based on the successful prior face detection. Other methods attack the more basic stage of the pipeline, such as Face Poison \cite{b22} and SSOD \cite{b23}. These methods primarily target the face detection and localization process, which serves as the foundational step in most deepfake retraining pipelines. By introducing adversarial perturbations specifically designed to disrupt the classification and bounding box regression of detectors, they can force the detection stage to fail or produce highly inaccurate coordinates. Such failures effectively terminate subsequent pre-processing steps like cropping and landmark alignment, thereby preventing attackers from acquiring usable facial data for model fine-tuning.

Interruption perturbations are an active defense strategy designed to distort the output of a deepfake model during the inference stage. These perturbations are often designed to be universal, meaning they are effective across different models and images. For instance, CMUA-Watermark \cite{b15} generates a cross-model universal adversarial watermark, which is achieved by optimizing the single perturbation pattern. When this perturbation pattern is added to the image, it can force deepfake models to produce an obvious distortion effect. Similarly, Tang et al. \cite{b20} assume that disrupting foundational features is key to a generalizable defense and specifically attacks an ensemble of common and pre-trained feature extractors. By destroying the fundamental semantic representations that most deepfake models rely on, their defense can effectively distort the quality of the final forged image across a wide range of different model architectures.

However, the preceding analysis shows that existing active defenses are often specialized. Interruption-based methods provide real-time protection but lack robustness against retraining, while poisoning-based methods can disrupt retraining but are ineffective against existing pre-trained deepfake models. This highlights the need for a more comprehensive framework. Our work TSDF is proposed to address this challenge by synergistically integrating these two distinct and defensive paradigms.

\subsection{Multifunctional Perturbation Methods}

To provide more comprehensive protection than single-function approaches, some studies have begun to explore multifunctional methods that combine different defensive goals. A common direction has been to merge interruption with traceability. For instance, Dual Defense \cite{b25} pioneers a dual-effect defense through a single embedded watermark that synergistically provides traceability while working to disrupt face-swapping models. Similarly, Zhu et al. \cite{b26} introduce an information-containing perturbation designed to combat facial manipulation systems by embedding an identity message, thus enabling traceability within the interrupted process. Wu et al. \cite{b27} propose a framework to resolve the common conflict between watermarking and detection. Their method fine-tunes a watermark, increasing a deepfake detector's accuracy on watermarked images without compromising traceability.

Although these advanced methods successfully demonstrate the value of multifunctionality, they share a common limitation regarding defense persistence. To solve this specific research gap, our Two-Stage Defense Framework pioneers the synergistic fusion of interruption and poisoning. Through a novel intensity separation mechanism, poisoning is clearly designed as a strategy to destroy the retraining process to counter adaptive attacks.

\begin{figure*}[htbp]
    \centering
    \includegraphics[scale=0.7]{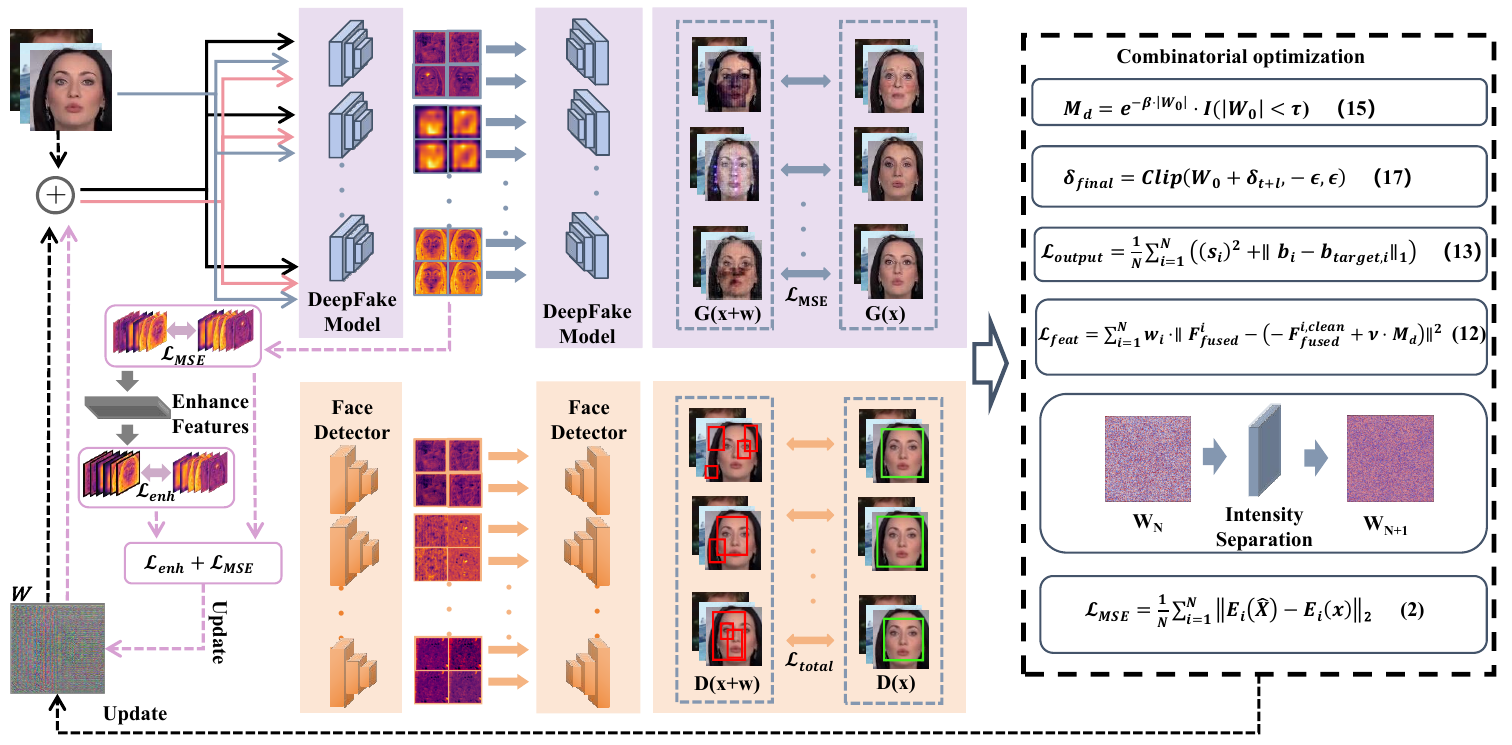}  
    \caption{The overall framework of the TSDF method. The process begins in the interruption stage (upper left), where an initial perturbation (W) is generated by feeding a perturbed image into several deepfake models and optimizing W to maximize the feature-level Mean Squared Error loss ($\mathcal{L}_{MSE}$) and minimizing a feature enhancement loss ($\mathcal{L}_{enh}$). Subsequently, the poisoning stage (lower left)  generates a specialized perturbation to attack multiple face detectors by maximizing feature-level loss ($\mathcal{L}_{feat}$) and output-level loss ($\mathcal{L}_{output}$). In the combination and optimization stage (right side), the two perturbations are efficiently fused. The combined system uses a threshold $\tau$ to identify low-intensity regions in the interruption perturbations, creating a mask where the poisoning perturbation is selectively applied. }
    \label{fig:03}
\end{figure*}

\section{METHOD}

This section details the TSDF. As illustrated in Fig. \ref{fig:03}, the framework is composed of three key components that work in sequence: interruption perturbation generation, poisoning perturbation generation and final fusion optimization. In the initial stage, a preliminary perturbation is designed to disrupt the feature extractors within deepfake models. This initial interruption perturbation then serves as a foundational component of the framework. It is utilized in the poisoning stage, where the perturbation targeting face detectors is precisely embedded into its low-intensity regions to create persistent protection. An optimization module fuses these two components into a unified perturbation that is visually imperceptible. The following subsections describe each component in detail.

\subsection{Deepfake Interruption via Perturbation}

Instead of directly attacking end-to-end deepfake models like CMUA \cite{b15}, the proposed method targets feature extractors used in deepfake networks. This prevents feature extractors from correctly encoding facial attributes, thereby disrupting downstream attribute editing and identity reconstruction. Given an input image x, an initial perturbation W is introduced to generate a perturbed image $\hat{X}$ = x+W. The goal is to maximize the discrepancy between extracted feature representations.

To achieve this, we optimize the initial perturbation to maximize the distance between extracted features from the original and perturbed images:
\begin{equation}
\max_W\sum_{i=1}^N\|E_i(\hat{X})-E_i(x)\|_2 ,
\end{equation}
where $E_i(\cdot)$ represents the feature extractor of the $i_{th}$ deepfake model, and $N$ is the number of targeted models. By disrupting multiple feature extractors simultaneously, the perturbation generalizes across different deepfake networks.

To train the perturbation, we define a loss function with two main components: a Mean Squared Error (MSE) loss and a feature difference enhancement loss. The MSE loss is used to quantify the difference between the feature representations of the original image x and the perturbed image $\hat{X}$:
\begin{equation}
\mathcal{L}_{MSE} = \frac{1}{N} \sum_{i=1}^{N} \| E_{i}(\hat{X}) - E_{i}(x) \|_2.
\end{equation}

This loss ensures that the perturbation creates a consistent and significant distortion across the feature extractors of all targeted models. To achieve a more potent disruption than a simple MSE loss can provide, we introduce a feature enhancement loss $\mathcal{L}_{enh}$ that amplifies discrepancies across three key levels: local, global, and structural. This amplification is driven by a non-linear exponential function, which is critical for magnifying even small feature differences into a significant and destructive effect.

\begin{equation}
\mathcal{F}_{enh} = E_i(\hat{X}) + \alpha\cdot e^{\frac{\|\hat{W}\circ\Delta\|}{\sigma}}\cdot\sum_{i=1}^3 w_i\Delta_i .
\label{eq:3}
\end{equation}

The core design of $\mathcal{F}_{enh}$ is to construct an amplified reference target in the feature space that the deepfake model must deviate from. Specifically, $\mathcal{F}_{enh}$ is constructed by taking the current perturbed feature $E_i(\hat{X})$ and shifting it along a direction defined by the discrepancy vector $\Delta$. This vector $\Delta$ aggregates local ($L_D$), global ($G_D$), and structural ($S_D$) differences between the perturbed and clean states. The exponential term $e^{\frac{||W\circ\Delta||}{\sigma}}$ functions as a non-linear gain, significantly magnifying even minute feature variations to create a state of maximized disruption. This approach ensures that the perturbation targets a wider range of semantic and structural components compared to standard linear metrics.

\begin{equation}
\mathrm{L_D} = Norm({E_i(\hat{X})}) - Norm({E_i(x)}) ,
\end{equation} 

\begin{equation}G_D=\frac{U(E_i(\hat{X}))-U(E_i(x))}{S(E_i(x))+z} ,
\end{equation}

\begin{equation}
\mathrm{S_D} = Att(Norm({E_i(\hat{X})})) - Att(Norm({E_i(x)})) ,
\end{equation}

\begin{equation}\hat{W}=\mathrm{Sigmoid}(U(|\Delta|)) ,
\end{equation}
where $Norm(\cdot)$ denotes normalization function to standardize the feature map and $U(\cdot)$ calculates the mean of a feature map across its spatial dimensions. $S (\cdot)$ calculates the standard deviation of a feature map. $Att (\cdot)$ is an attention mechanism applied to normalized feature maps to compute and weight their structural feature. $z$ is a numerical stability constant.

Consequently, the feature enhancement loss $\mathcal{L}_{enh}$ is formulated to align the current adversarial feature $E_i(\hat{X})$ with the amplified target $\mathcal{F}_{enh}$:
\begin{equation}
\mathcal{L}_{enh} = \frac{1}{N} \sum_{i=1}^{N} \| \mathcal{F}_{enh} - E_{i}(\hat{X}) \|_2.
\end{equation}

By minimizing this specific loss, we force the perturbation to evolve towards larger and more diverse feature discrepancies than a standard MSE loss would achieve, ultimately resulting in a more destructive and generalizable interruption effect.

The total loss function is a weighted sum of these two components, balanced by a hyperparameter $\lambda$:
\begin{equation}\mathcal{L}_{\mathrm{feature}}=\lambda\mathcal{L}_{enh}+\mathcal{L}_{MSE}.
\end{equation}

To maximize the feature loss, we iteratively optimize the perturbation W using gradient ascent. This optimization follows the Momentum Iterative Fast Gradient Sign Method (MI-FGSM) \cite{b28}, Unlike standard methods, MI-FGSM integrates a momentum term to stabilize the update direction. The optimization step is formulated as:
\begin{equation} W_{t+1}=\mathrm{Proj}_{\epsilon}(W_t+\gamma\cdot\mathrm{sign}(g_{t+1})),\label{eq:6} \end{equation}
where $\gamma$ is the step size and $\epsilon$ is the perturbation budget. The projection operator ensures that the interruption perturbation remains within a bounded perturbation space, and $g_{t+1}$ denotes the accumulated gradient with momentum factor $\mu$.

\subsection{Poisoning the Retraining Pipeline}

While interruption is effective for real-time image distortion, long-term defense must prevent an attacker from using perturbed examples to fine-tune their model. Our approach creates an effective perturbation designed to disrupt face detection, which is the crucial first step in an attacker's data processing pipeline.

To ensure a generalizable poisoning effect against a range of detectors, we employ a joint attack strategy at the feature-level. This principle of attacking across the feature layers of multiple networks has also been explored in methods like Face Poison \cite{b22}. As formulated in Equation \ref{eq:7}, our approach creates a fused feature representation by combining outputs from the feature layers of multiple state-of-the-art (SOTA) detectors. The feature-level loss $\mathcal{L}_{feat}$ is designed to maximize the difference between features within the masked poisoning regions ($M_d$). For a batch of images, the fusion features are:
\begin{equation}
F_{\text{fused}} = \sum_{i=1}^{N} \alpha_i \cdot D_i(\text{x}) ,
\label{eq:7}
\end{equation}
\begin{equation}
    \mathcal{L}_{feat} = \sum_{i=1}^{N} w_i \cdot \|F_{\text{fused}}^i - (-F_{\text{fused}}^{i,\text{clean}} + \nu \cdot M_d)\|^2 ,
\end{equation}
where $D_i(\cdot)$ is the feature extracted from the face detector $i_{th}$. $\alpha_i$ is the fusion weights that balance the contribution of each of the detectors and $w_i$ is the weight of the layer $i_{th}$. $N$ is the number of feature layers to attack. $\nu$ is a hyperparameter that controls the magnitude of the targeted feature-space margin. $F_{\text{fused}}^{i,\text{clean}}$ denotes the fused feature representation derived from the original image. The optimization of this poisoning loss is limited to specific regions $M_d$ defined by our intensity separation mechanism, and its technical details are given in the following section (III-C).

In addition to the feature-level loss, the framework employs a more direct attack on the detector's final predictions through the output-level loss. This loss is composed of two parts: a score loss, which pushes the confidence scores of all detected boxes to zero, and the box loss minimizes the size of the predicted bounding boxes. The total output loss is a weighted sum of these components:
\begin{equation}\mathcal{L}_{output}=\frac{1}{N}\sum_{i=1}^N\left((s_i)^2+\|b_i-b_{target,i}\|_1\right) ,
\end{equation} 
where $s_i$ is the confidence score of the $i_{th}$ detected box. $||b_i-b_{target,i}\|_1$ is the sum of the absolute differences between the coordinates of the predicted box and the target box that forces the predicted box to shrink to a minimal size.
\begin{equation}\mathcal{L}_{total}=\mathcal{L}_{feat}-\mathcal{L}_{output} .
\label{eq:14} 
\end{equation}

The $\mathcal{L}_{total}$ constructs a multi-objective optimization that simultaneously disrupts feature-layer representations ($\mathcal{L}_{feat}$) and detector-level outputs ($\mathcal{L}_{output}$). This dual-level approach constrains both feature consistency and bounding box geometry, thereby increasing the difficulty of model adaptation during retraining. 
The poisoning perturbation is then optimized to maximize this total loss $\mathcal{L}_{total}$, which effectively disrupts face detection. The optimization remains constrained to the masked regions to maintain image quality.

\subsection{Perturbation Fusion Optimization}
\begin{figure}[htbp]
    \centering
    \includegraphics[scale=0.55]{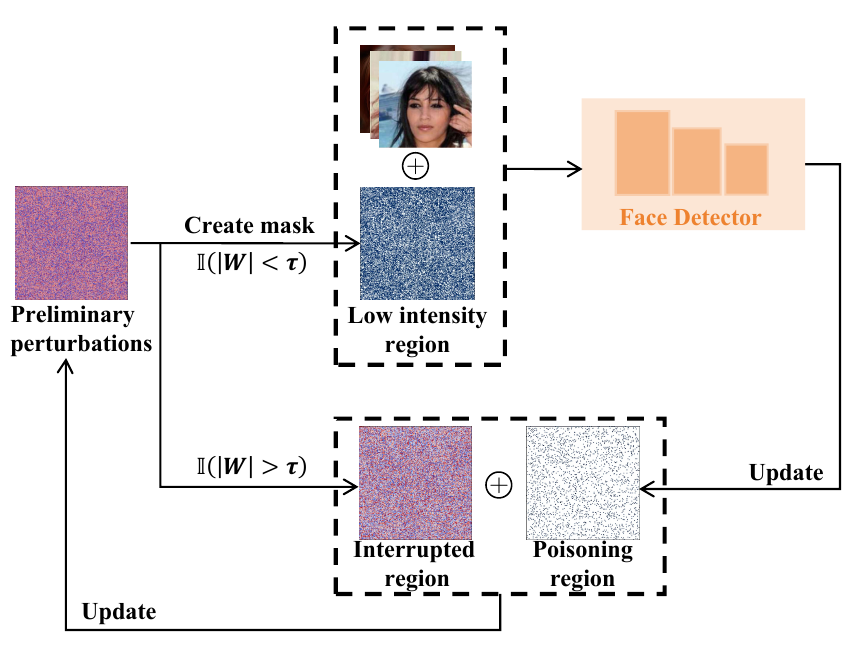}  
    \caption{Implementation diagram of intensity separation mechanism.}
    \label{fig:02}
\end{figure}

The interruption perturbation and the subsequent poisoning perturbation are then synergistically combined into a unified perturbation. Directly combining the two defensive goals into a single weighted-sum loss function is often suboptimal, as competing gradients can lead to functional conflicts \cite{b18}. To avoid this, our intensity separation mechanism decouples the two tasks spatially by utilizing the low-intensity regions of the primary interruption perturbation. This strategy allows both the interruption and poisoning functions to operate effectively without mutual interference, ensuring a more effective and efficient unified perturbation. The process of this mechanism is illustrated in Fig. \ref{fig:02}.

To avoid the challenges of a direct multi-objective optimization via a single loss function, this mechanism employs a region-based fusion strategy. It works by first identifying the regions within the initial interruption perturbation $W_0$, where the signal magnitude is low. The poisoning mask $M_d$, is then generated to isolate these low-intensity areas, designating them as the channels for the poisoning component. Furthermore, an attenuation coefficient $\beta$ scales the poisoning perturbation's strength, ensuring it is applied precisely where it will not interfere with the high-intensity components of the interruption.
we compute the magnitude map: 
 
\begin{equation}
M_d = e^{-\beta \cdot |W_0|} \cdot I({|W_0| < \tau}) ,
\label{eq:15}
\end{equation}
where $\tau$ is a predefined threshold and the exponential term $e^{-\beta\cdot|W_0|}$ functions as a principled soft-assignment strategy for adversarial budget allocation. Theoretically, the magnitude of $|W_0|$ represents the local sensitivity required to maintain the interruption effect. By employing exponential decay, the framework ensures that poisoning energy is adaptively concentrated in regions with minimal interruption signals, effectively navigating the trade-off between the two defensive functions. Compared to a rigid binary mask, this continuous scaling offers a more fine-grained optimization landscape within the region. By assigning weights based on magnitude rather than a uniform binary value, it provides informative gradients that guide the optimizer to prioritize less sensitive regions. This mechanism mitigates the optimization rigidity often associated with discrete masking, reducing potential functional conflicts during the iterative fusion process.

The poisoning perturbation $\delta$ is then optimized within these masked regions $M_d$. Since this stage requires precise convergence within the residuals, we employ a constrained iterative optimization strategy. The update rule is formulated as:
\begin{equation}
\delta_{t+1} = \text{Clip}(\delta_t - \eta \cdot \nabla_{\delta}L_{\text{total}}, -\epsilon, \epsilon) \odot M_d ,
\end{equation}

\begin{equation}
\delta_{\text{final}} = \text{Clip}(W_0 + \delta_{\text{t+1}}, -\epsilon, \epsilon) ,
\end{equation}
where $\eta$ is the step size and $\odot$ denotes element-wise multiplication. This fusion strategy ultimately produces a single perturbation with dual defensive capabilities. The complete workflow of our proposed framework is summarized in Algorithm \ref{alg:tsdf_formatted}.

\begin{algorithm}[h!]
\caption{Two-Stage Defense Framework (TSDF)}
\label{alg:tsdf_formatted}
\begin{algorithmic}[1]
    \STATE \textbf{Input:} image $x$, budget $\epsilon$, threshold $\tau$, initial perturbation W, poisoning perturbation $\delta$, iterations $T_{int}, T_{poi}$.
    \STATE \textbf{Output:} optimized perturbation $\delta_{final}$.
    
    \STATE \textit{// Stage 1: Generate Interruption Perturbation.}
    \STATE Initialize interruption perturbation $W_0$ randomly.
    \FOR{$t = 1$ \textbf{to} $T_{int}$}
        \STATE Update W using $\mathcal{L}_{\mathrm{feature}}$ (Eq. 9) via MI-FGSM to maximize.
    \STATE Set $W_0 \leftarrow W_{T_{int}}$.
    \ENDFOR
    \STATE \textit{// Stage 2: Generate and Fuse Poisoning Perturbation.}
    \STATE Initialize poisoning perturbation $\delta_{0} \leftarrow \mathbf{0}$.
    \STATE Create poisoning mask $M_d \leftarrow e^{-\beta \cdot |W_0|} \cdot \mathbf{I}(|W_0| < \tau)$.
    \FOR{$t = 1$ \textbf{to} $T_{poi}$}
        \STATE Update $\delta$ using
        $\mathcal{L}_{\mathrm{total}}$ (Eq. 14) and the mask $M_d$.
    \STATE Set $\delta_{poison} \leftarrow \delta_{T_{poi}}$.
    \STATE Combine perturbations: $\delta_{final} \leftarrow W_0 + \delta_{poison}$.
    \ENDFOR
    \RETURN $\text{Clip}(\delta_{final}, -\epsilon, \epsilon)$.
\end{algorithmic}
\end{algorithm}

\begin{table*}[htbp]
    \centering
    \caption{Comparison of Interruption Performance among CMUA, FOUND and TSDF. The avg is the average result of each indicator under this dataset. The best result is marked in bold.}
    \scriptsize
    \setlength{\tabcolsep}{3pt}
    \resizebox{\textwidth}{!}{
        \begin{tabular}{@{}llcccccccccccccccc@{}}
            \toprule
            \multirow{2}{*}{Dataset} & \multirow{2}{*}{Deepfake Model} & \multicolumn{3}{c}{L2mask$\uparrow$} & \multicolumn{3}{c}{SRmask$\uparrow$} & \multicolumn{3}{c}{FID$\uparrow$} & \multicolumn{3}{c}{PSNR$\downarrow$} & \multicolumn{3}{c}{SSIM$\downarrow$} \\
            \cmidrule(lr){3-5} \cmidrule(lr){6-8} \cmidrule(lr){9-11} \cmidrule(lr){12-14} \cmidrule(lr){15-17}
            & & CMUA & FOUND & TSDF & CMUA & FOUND & TSDF & CMUA & FOUND & TSDF & CMUA & FOUND & TSDF & CMUA & FOUND & TSDF \\
            \midrule
            \multirow{5}{*}{CelebA} 
            & StarGAN & 0.21 & 0.26 & \textbf{0.37} & \ 100.00\% & 100.00\% & \textbf{100.00\%} & 305.05 & 362.54 & \textbf{378.89} & 12.85 & 11.45 & \textbf{9.73} & 0.48 & 0.26 & \textbf{0.17} \\
            & AGGAN    & \textbf{0.23} & 0.19 & 0.18 & 99.93\% & 100.00\% & \textbf{100.00\%} & 163.00 & \textbf{224.97} & 221.36 & \textbf{15.93} & 16.12 & 17.86 & \textbf{0.62} & 0.68 & 0.70 \\
            & AttGAN   & 0.13 & 0.15 & \textbf{0.17} & 87.01\% & 97.90\% & \textbf{98.25\%} & 180.95 & 183.22 & \textbf{192.31} & 18.59 & 17.36 & \textbf{16.76} & 0.69 & 0.64 & \textbf{0.62} \\
            & HiSD     & \textbf{0.21} & 0.09 & 0.17 & \textbf{99.72\%} & 92.55\% & 99.15\% & \textbf{188.85} & 183.46 & 170.99  & \textbf{16.09} & 19.11 & 16.14 & \textbf{0.75} & 0.82 & 0.76 \\
            \cmidrule[0.4pt]{2-17}
            & Avg      & 0.20 & 0.17 & \textbf{0.22} & 96.67\% & 97.61\% & \textbf{99.35\%} & 209.46 & 238.55 & \textbf{240.88} & 15.87 & 16.01 & \textbf{15.12} & 0.64 & 0.60 & \textbf{0.56} \\
            \midrule
            \multirow{5}{*}{LFW} 
            & StarGAN & 0.21 & 0.26 & \textbf{0.36} & \ 100.00\% & 100.00\% & \textbf{100.00\%} & 298.86 & \textbf{398.86} & 372.80 & 12.92 & 15.86 & \textbf{10.32} & 0.47 & 0.27 & \textbf{0.20} \\
            & AGGAN    & \textbf{0.24} & 0.20 & 0.19 & \ 100.00\% & 100.00\% & \textbf{100.00\%} & 193.57 & \textbf{269.06} & 228.19 & 16.10 & \textbf{15.72} & 17.97 & \textbf{0.61} & 0.66 & 0.69 \\
            & AttGAN   & 0.11 & 0.16 & \textbf{0.16} & 95.24\% & 98.05\% & \textbf{99.01\%} & 207.76 & 248.99 & \textbf{253.76} & 18.51 & 16.70 & \textbf{16.17} & 0.67 & 0.60 & \textbf{0.59} \\
            & HiSD     & 0.12 & 0.07 & \textbf{0.12} & \textbf{97.44\%} & 88.04\% & 95.42\% & \textbf{221.82} & 163.23 & 188.98 & \textbf{16.22} & 19.61 & 16.58 & \textbf{0.75} & 0.82 & 0.76 \\
            \cmidrule[0.4pt]{2-17}
            & Avg      & 0.17 & 0.17 & \textbf{0.21} & 98.17\% & 96.52\% & \textbf{98.61\%} & 230.50 & \textbf{270.04} & 260.93 & 15.84 & 16.97 & \textbf{15.26} & 0.63 & 0.59 & \textbf{0.56} \\
            \midrule
            \multirow{5}{*}{FF++O} 
            & StarGAN & 0.21 & 0.24 & \textbf{0.35} & 100.00\% & 100.00\% & \textbf{100.00\%} & 377.31 & 390.27 & \textbf{400.15} & 13.58 & \textbf{8.49} & 10.74 & 0.55 & 0.37 & \textbf{0.23} \\
            & AGGAN    & \textbf{0.24} & 0.22 & 0.19 & 100.00\% & 100.00\% & \textbf{100.00\%} & 276.09 & \textbf{370.10} & 299.18 & 15.14 & \textbf{13.25} & 17.66 & \textbf{0.70} & 0.77 & 0.74 \\
            & AttGAN   & 0.14 & \textbf{0.21} & 0.16 & 94.80\% & \textbf{99.51\%} & 99.02\% & 216.97 & \textbf{282.66} & 260.94 & 15.71 & 16.81 & \textbf{15.47} & 0.70 & 0.60 & \textbf{0.57} \\
            & HiSD     & 0.11 & 0.09 & \textbf{0.11} & 86.15\% & 85.35\% & \textbf{94.95\%} & \textbf{213.57} & 178.71 & 176.98 & 17.99 & 17.10 & \textbf{16.67} & 0.83 & 0.88 & \textbf{0.81} \\
            \cmidrule[0.4pt]{2-17}
            & Avg      & 0.18 & 0.19 & \textbf{0.20} & 95.24\% & 96.22\% & \textbf{98.49\%} & 270.99 & \textbf{305.44} & 284.31 & 15.61 & \textbf{13.91} & 15.14 & 0.70 & 0.66 & \textbf{0.59} \\
            \bottomrule
        \end{tabular}
        \label{tab:table2}
    }
\end{table*}

\begin{table*}[!ht]
    \centering
    \caption{Comparison of Face Detector Performance Under Various Poisoning Attacks.}
    \resizebox{\textwidth}{!}{
        \begin{tabular}{ll*{9}{c}}
            \toprule
            \multirow{2}{*}{Metric} & \multirow{2}{*}{Method} & \multicolumn{3}{c}{CelebA} & \multicolumn{3}{c}{LFW} & \multicolumn{3}{c}{FF++O} \\
            \cmidrule(lr){3-5} \cmidrule(lr){6-8} \cmidrule(lr){9-11}
            & & DSFD & RetinaFace & S3FD & DSFD & RetinaFace & S3FD & DSFD & RetinaFace & S3FD \\
            \midrule
            \multirow{5}{*}{F1-score$\downarrow$} 
            & Original   & 0.99   & 0.99   & 1.00  & 0.99   & 0.98   & 1.00   & 0.99   & 0.99   & 1.00  \\
            & Random     & 0.99   & 0.85   & 0.99   & 0.99   & 0.85   & 0.99   & 0.99   & 0.84   & 0.99   \\
            & SSOD       & 0.81   & 0.90   & 0.99   & 0.81  & 0.96   & 0.97   & 0.80   & 0.92   & 0.98   \\
            & Face Poison& 0.86   & 0.96   & 0.96   & 0.85  & 0.97   & 0.96   & 0.86   & 0.94   & 0.95   \\
            & TSDF       & \textbf{0.29}   & \textbf{0.77}   & \textbf{0.23}   & \textbf{0.43}  & \textbf{0.79}   & \textbf{0.31}   & \textbf{0.47}   & \textbf{0.79}   & \textbf{0.25}   \\
            \bottomrule
        \end{tabular}
    }
    \label{tab:table3}
\end{table*}

\section{EXPERIMENTS}

\subsection{Experimental Setup}
Datasets: We conduct the experiments on the CelebA dataset \cite{b31} for training. The dataset comprises over 200,000 face images, from which we use the aligned and cropped version. To ensure a fair comparison with the SOTA baseline FOUND \cite{b20}, we adopt a training set of 128 images. As shown in FOUND, the perturbation performance saturates at this sample size meaning 128 images are sufficient to capture the common features for constructing a universal attack. For evaluation, we then apply this generated perturbation to all images in the official test split. To further evaluate the generalization capability of TSDF, additional datasets are introduced, including the Labeled Faces in the Wild (LFW) dataset \cite{b32} and facial images extracted from the FaceForensics++ Original video dataset (FF++O) \cite{b33}.  

Deepfake Model: We evaluate the defense against four deepfake models, HiSD \cite{b34}, AttentionGAN (AGGAN) \cite{b35}, StarGAN \cite{b36}, and AttGAN \cite{b37}. For StarGAN and AGGAN, we select 5 attributes: black hair, blonde hair, brown hair, gender, and age. In AttGAN's training, the editing tasks are expanded to 14 attributes. Given HiSD’s advanced facial editing capabilities, the experiments incorporate modifications of the glasses attribute. To ensure experimental consistency and fairness, all deepfake models are trained on the CelebA dataset.

Face Detector: To evaluate the effectiveness of TSDF in data poisoning scenarios, we select several widely used face detection models based on Deep Neural Networks (DNNs). Specifically, we employ the VGG16 network integrated with DSFD \cite{b21} and S3FD \cite{b30}, the RetinaFace \cite{b38} based on MobileNet-0.25. Additionally, all face detection models are trained by default under their own conditions.

Baseline: To rigorously evaluate our framework, our baselines are selected to represent the most foundational, representative and directly competing SOTA methods. For interruption, we compare with the pioneering universal defense method CMUA-Watermark \cite{b15} and select FOUND \cite{b20} as a critical baseline. The latter's principle of attacking an ensemble of feature extractors is similar to our own interruption module, making it the most relevant SOTA benchmark to validate our feature enhancement part. For the poisoning task, we benchmark against Face Poison \cite{b22}, the most direct contemporary competitor that also targets face detection, and the foundational SSOD \cite{b23} which provides a representative baseline for disrupting the fundamental classification and localization mechanisms within the face detection pipeline to validate our effectiveness against detector-level attack strategies. Given the limited availability of public code, these poisoning baselines are rigorously re-implemented from their original papers to ensure a fair comparison.

Implementation Details: The hyperparameter configuration is aligned with existing baseline methods while jointly optimizing the interruption and poisoning effects of the perturbations. The perturbation range for both attacks is set to (0.05, 0.05) to balance visual fidelity with attack efficacy. In the training stage, 50 and 30 iterations are performed for interruption and poisoning, as we observe in preliminary experiments that performance is saturated beyond these points. For the optimizer parameters in Equation \ref{eq:6} and \ref{eq:15}, $\lambda$ and $\beta$ are set to 0.001 and 0.4, respectively. The intensity separation threshold $\tau$ is set to 0.04, a value determined to provide the optimal balance between defensive functions, as validated in the ablation study. For the interruption stage utilizing MI-FGSM, the momentum decay factor $\mu$ is set to 0.8.
\subsection{Evaluation Metrics}

Interruption Metrics: We use L2mask value and SRmask \cite{b20} to measure the success of the interruption. A higher L2mask value indicates a more successful disruption. The Fréchet Inception Distance (FID) is used to quantify the perceptual difference between the original images and the distorted outputs, with higher values indicating a stronger interruption effect. To assess the disruptive effect, we also employ the Peak Signal-to-Noise Ratio (PSNR) and the Structural Similarity Index Measure(SSIM). Both metrics are used to measure the discrepancy between the original images and the distorted output, where lower scores indicate a better interruption.

Poisoning Metric: We use the F1-score \cite{b22} to measure the attack's effectiveness against face detectors. The F1-score assesses the detector's precision and recall. A lower F1-score reflects a significant degradation in the detector's performance, which directly corresponds to a more successful poisoning attack.

Image Quality Metrics: We use PSNR and SSIM to measure the visual quality of the protected images compared to the originals. Higher PSNR and SSIM values indicate that the perturbation is imperceptible. PSNR and SSIM are used in two distinct contexts: for evaluating interruption, a lower score between the forged and original images indicates a more effective defense. In contrast, to assessing image quality, a higher score indicates better visual imperceptibility.

\subsection{Compare the SOTA Methods}
This section validates the effectiveness of the proposed TSDF method through a series of direct comparisons with SOTA methods. 

\paragraph{Interruption Performance Against SOTA Methods}
We evaluate the interruption capability of TSDF against SOTA methods, CMUA-Watermark \cite{b15} and FOUND \cite{b20}, with detailed results presented in Table \ref{tab:table2}. The results show that TSDF consistently delivers a more powerful and destructive effect on the final forged images across multiple datasets and models. A key advantage of the method is its superior ability to degrade the structural integrity of the forged output. This is most obvious in SSIM scores. On the CelebA dataset, TSDF achieves an average SSIM of 0.56, which is significantly lower than the scores of CMUA and FOUND, indicating superior disruption. This trend is particularly strong against models like StarGAN, where TSDF achieves a remarkable SSIM of just 0.17, far exceeding the disruptive effect of baseline. Furthermore, this disruptive force is reflected in other key metrics. On the LFW dataset, TSDF achieves the highest average L2mask of 0.21 and the highest average SRmask of 98.61\%, indicating the most powerful disruption among the three methods. The minor difference in some of TSDF's initial interruption metrics is an expected outcome of our dual-function design. Unlike single-function baselines that dedicate their entire budget to interruption, our framework must balance both interruption and poisoning capabilities within the same perturbation. Our ablation study quantitatively demonstrates that this balance is a deliberate design choice, made to achieve long-term persistence.

\begin{table*}[htbp]
\centering
\caption{Degradation of Baseline Interruption Performance After Retraining.}
\begin{tabular}{lc*{2}{c}*{2}{c}*{2}{c}*{2}{c}*{2}{c}}
\toprule
\multirow{2}{*}{Method} & \multirow{2}{*}{Deepfake Model} & \multicolumn{2}{c}{L2mask↑} & \multicolumn{2}{c}{SRmask↑} & \multicolumn{2}{c}{FID↑} & \multicolumn{2}{c}{PSNR↓} & \multicolumn{2}{c}{SSIM↓} \\
\cmidrule(lr){3-4} \cmidrule(lr){5-6} \cmidrule(lr){7-8} \cmidrule(lr){9-10} \cmidrule(l){11-12}
 & & Original & Retrained & Original & Retrained & Original & Retrained & Original & Retrained & Original & Retrained \\
\midrule
\multirow{5}{*}{CMUA} 
 & StarGAN & 0.21 & 0.07 & 100.00\% & 71.15\%  & 305.05  & 131.62  & 12.85   & 17.86   & 0.48 & 0.70  \\
 & AGGAN   & 0.23 & 0.08 & 99.93\%  & 95.20\%   & 163.00     & 83.59   & 15.93   & 23.09   & 0.62 & 0.78 \\
 & AttGAN  & 0.13 & 0.07 & 87.01\%  & 55.22\%   & 180.95  & 113.3   & 18.59   & 24.43   & 0.69 & 0.83 \\
 & HiSD    & 0.21 & 0.11 & 99.72\%  & 94.75\%  & 188.85  & 181.18  & 16.09   & 17.07   & 0.75 & 0.83 \\
 \cmidrule[0.4pt]{2-12}
 & Avg     & 0.20 & 0.08 & 96.66\% & 79.08\% & 209.46 & 127.42 & 15.87 & 20.61 & 0.64 & 0.79 \\
\midrule
\multirow{5}{*}{FOUND} 
 & StarGAN & 0.26 & 0.21 & 100.00\% & 91.03\%  & 362.54  & 298.86  & 11.45   & 12.92   & 0.26 & 0.47 \\
 & AGGAN   & 0.19 & 0.05 & 100.00\% & 46.11\%  & 224.97  & 193.57  & 16.12   & 24.54   & 0.68 & 0.83 \\
 & AttGAN  & 0.15 & 0.09 & 97.90\%   & 63.24\%  & 183.22  & 117.76  & 17.36   & 23.13   & 0.64 & 0.82 \\
 & HiSD    & 0.09 & 0.07 & 92.55\%  & 81.24\%  & 183.46  & 141.82  & 19.11   & 19.96   & 0.82 & 0.84 \\
 \cmidrule[0.4pt]{2-12}
 & Avg     & 0.17 & 0.11 & 97.61\% & 70.41\% & 238.55    & 188.00 & 16.01   & 18.67   & 0.60  & 0.74  \\
\bottomrule
\end{tabular}
\label{tab:table4}
\end{table*}

\paragraph{Poisoning Performance Against SOTA Methods}
We then assess the effectiveness of TSDF's poisoning component against face detectors, comparing it to random noise and reproduced versions of SSOD and FacePoison.
The results in Table \ref{tab:table3} are critical as they validate the core defense strategy. Instead of merely trying to survive an attacker's adaptation, our framework fundamentally disrupts the retraining process. By significantly affecting the automated face detectors that an attacker relies on, our method makes large-scale and automated collection of training data impractical. This represents a fundamental shift in defensive strategy. The objective is no longer to create a perturbation that passively withstands retraining, but to design one that actively destroys the retraining model. This capability is crucial for achieving more persistent defense.
For example, on the CelebA dataset, TSDF drastically reduces the F1-score for the RetinaFace detector to 0.77. This represents a far more successful attack compared to FacePoison and SSOD, which only achieve scores of 0.96 and 0.90 respectively. The advantage is even more pronounced against the S3FD detector, where our method achieves F1-score of 0.23, while the other methods barely impact the detector's performance. This superior performance is consistent across the LFW and FF++O datasets, where TSDF consistently achieves the lowest F1-score.
These results validate that our framework's poisoning component poses a significant threat to the face detection process. This capability is crucial for our two-stage defense, as it is the mechanism designed to prevent attackers from bypassing the interruption defense.

\begin{table}[h]
\centering
\caption{Comparative Analysis of Defense Persistence After Retraining. The table compares the interruption effectiveness of CMUA, FOUND, and the TSDF before and after a simulated adaptive attack. A lower SSIM or a higher FID indicates a more effective and persistent interruption defense.}
\setlength{\tabcolsep}{13pt}
\begin{tabular}{ccccc}
\toprule
\multirow{2}{*}{Method} & \multicolumn{2}{c}{SSIM↓} & \multicolumn{2}{c}{FID↑} \\
\cmidrule(lr){2-3} \cmidrule(lr){4-5} 
& Before & After & Before & After \\
\midrule
CMUA & 0.59 & 0.77 & 243.00 & 122.46 \\
\midrule
FOUND & 0.45 & 0.65 & 272.88 & 253.31 \\
\midrule
TSDF & 0.40 & 0.26 & 285.56 & 344.12 \\
\bottomrule
\end{tabular}
\label{tab:table9}
\end{table}

\begin{table}[h]
\centering
\caption{Model Performance discrepancy After Retraining on TSDF Data. This table compares the performance of the original baseline model (Healthy) against the same model after it has been retrained on TSDF data (Poisoned).}
\setlength{\tabcolsep}{15pt}
\renewcommand{\arraystretch}{1.5}
\begin{tabular}{llcc}
\toprule
Deepfake Model & Model State & FID$\uparrow$ & SSIM$\downarrow$ \\
\midrule
\multirow{2}{*}{StarGAN} & Healthy & 43.77 & 0.72 \\
                         & Poisoned & 199.87 & 0.26 \\
\midrule
\multirow{2}{*}{AttGAN}  & Healthy & 43.03 & 0.89 \\
                         & Poisoned & 121.06 & 0.68 \\
\bottomrule
\end{tabular}
\label{tab:table8}
\end{table}

\begin{figure*}[htbp]
    \centering
    \includegraphics[width=\textwidth]{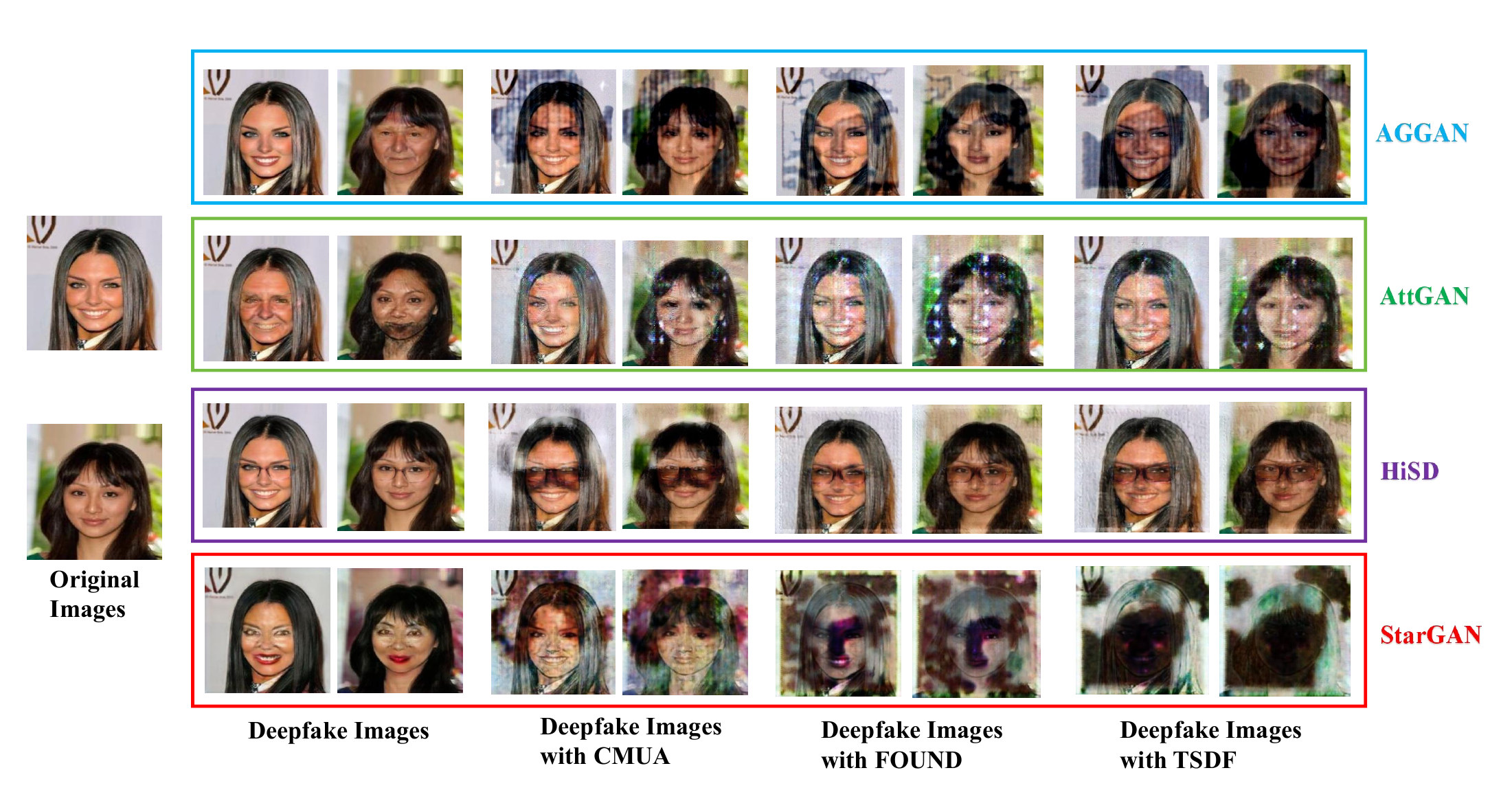}  
    \caption{Visual comparison of the interruption effect. This figure compares the outputs of deepfake models when protected by CMUA, FOUND, and TSDF.}
    \label{fig:04}
\end{figure*}

\begin{figure}[htbp]
    \centering
    \includegraphics[scale=0.36]{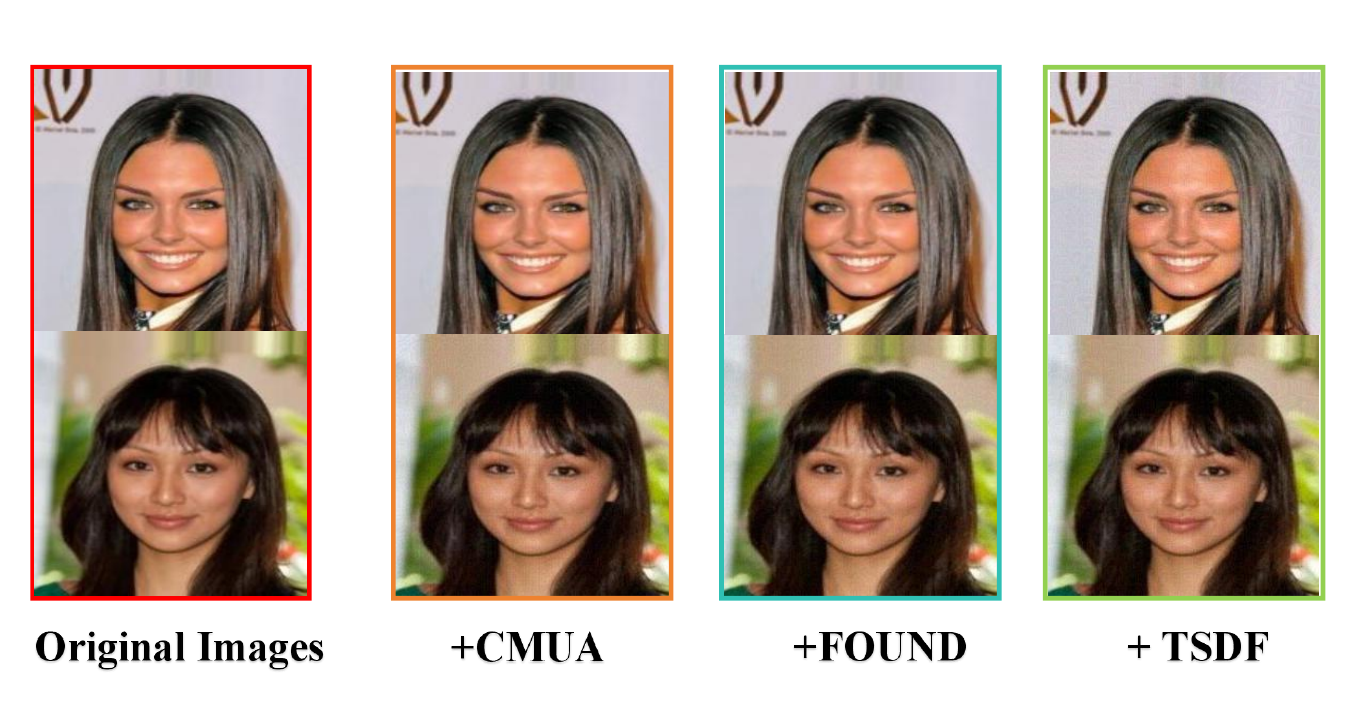}  
    \caption{Comparison of image quality of different methods.}
    \label{fig:05}
\end{figure}

\begin{table}[htbp]
    \centering
     \caption{Comparison of SSIM and PSNR indicators on different datasets.}
    \setlength{\tabcolsep}{6pt}
    \renewcommand{\arraystretch}{1.5}
    \begin{tabular}{lcccccc}
        \hline
        \multirow{2}{*}{Dataset} 
        & \multicolumn{3}{c}{SSIM$\uparrow$} 
        & \multicolumn{3}{c}{PSNR$\uparrow$} \\
        \cline{2-4} \cline{5-7}
        & CMUA & FOUND & TSDF 
        & CMUA & FOUND & TSDF \\
        \hline
        CelebA 
        & 0.88   & 0.89 & \textbf{0.89} 
        & 32.45 & 33.03 & \textbf{33.07} \\
        LFW    
        & 0.88   & 0.89 & \textbf{0.89} 
        & 32.49 & 33.07 & \textbf{33.09} \\
        FF++O  
        & 0.87   & 0.88 & \textbf{0.89} 
        & 32.48 & 33.06 & \textbf{33.08} \\
        \hline
    \end{tabular}
    \label{tab:table5}
\end{table}

\subsection{Analysis of Defense Persistence}
This section is dedicated to one of the contributions of this work, which is to validate the persistence of our framework. We conduct a direct comparative analysis that first establishes the failure of conventional SOTA defenses against an adaptive attack. We then demonstrate that our framework not only withstands the same attack but also enhances its defensive capabilities, thereby proving its superior persistence.

\paragraph{Baseline Performance under Adaptive Attack}
A core motivation for this work is to address the lack of persistence in conventional active defense methods. To quantitatively demonstrate this vulnerability, we simulate a powerful adaptive attack scenario designed to test the limits of existing interruption-only defenses. 

To simulate a worst-case attack scenario, we create training datasets composed entirely of protected images. Specifically, we apply the SOTA interruption perturbations from CMUA and FOUND to the CelebA training set, creating two separate protected datasets. The deepfake models are then retrained on these respective datasets.  

The results presented in Table \ref{tab:table4} reveal a critical vulnerability in conventional interruption-only methods. For instance, with the CMUA method, the average SSIM score on the CelebA dataset increases from 0.64 to 0.79 after retraining, while the average FID score falls from 209.46 to 127.42. The rising SSIM indicates that the output of perturbed examples has become more similar to the original images, while the falling FID confirms they have become more realistic and less distorted. Both metrics prove that the interruption effect has been largely nullified. This degradation provides clear quantitative evidence that adaptive attacks severely compromise conventional interruption-only defenses, proving their lack of persistence. This highlights the key vulnerability our TSDF method is designed to solve.

\paragraph{Comparative Analysis of TSDF's Persistence}
To provide a direct comparison against this established baseline of failure, we subjected our TSDF method to the identical adaptive attack.

To validate the persistence of our framework, we conduct a direct comparative experiment against the SOTA interruption methods. To ensure a fair comparison, we evaluate all methods on the same two representative models, StarGAN and AttGAN. We simulate an adaptive attack by retraining these models on datasets protected by each respective defense. For TSDF, the retraining dataset is first processed by the poisoning mechanism. We then measure the interruption effectiveness of each defense before and after this adaptive attack to quantify their persistence.

The results presented in Table \ref{tab:table9} provide clear quantitative evidence of TSDF's superior persistence. After retraining, the defenses of both CMUA and FOUND degrade significantly. CMUA's SSIM score increase from 0.59 to 0.77, and its FID score drops from 243.00 to 122.46. These results confirm that the retrained models successfully learn to bypass these perturbations. In contrast, the output from the model retrained on TSDF-protected data become even more distorted. Its SSIM score decreases from 0.40 to 0.26, and its FID score increases from 285.56 to 344.12. This outcome is not because the interruption perturbation became stronger. It is direct evidence of the poisoning component's success in degrading the quality of deepfake models.

This enhanced persistence is attributed to the efficacy of the poisoning component. As detailed in the model discrepancy analysis in Table \ref{tab:table8}, retraining on TSDF data fundamentally destroys the deepfake model's own generative capabilities. For instance, a StarGAN model's FID score on clean inputs after this process increases from a baseline of 43.77 to 199.87. Since the model's core ability to generate quality images is compromised, it cannot learn to properly adapt to or bypass the complex interruption perturbation.

This shows the important synergy of our framework. The poisoning component does not merely protect the interruption defense, but turns the attacker's own retraining attempt into a self-destructive process that destroys their deepfake model.

\subsection{Visual and Qualitative Analysis}
The quantitative results are supported by the qualitative analysis. It confirms the framework's success by showing that the final perturbation is nearly imperceptible while still causing both highly distorted deepfake outputs and a failure of face detectors.

\begin{figure}[htbp]
    \centering
    \includegraphics[scale=0.53]{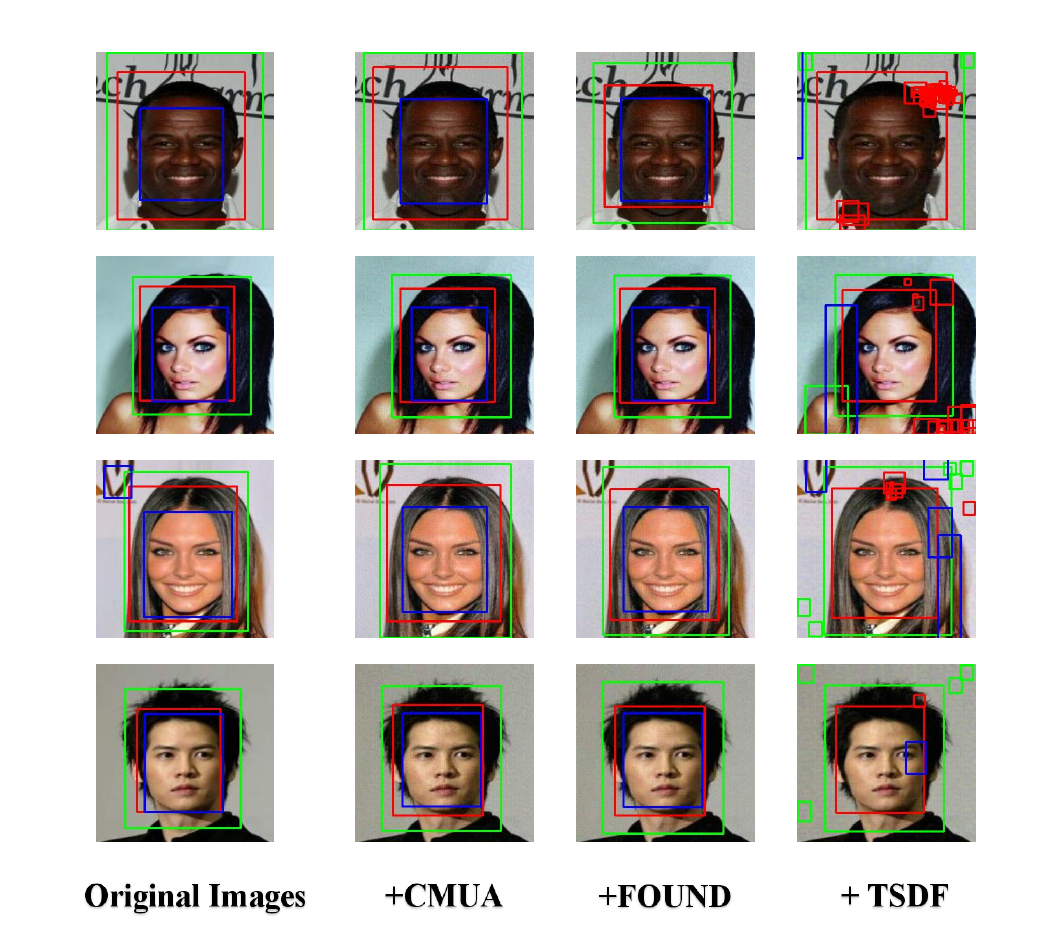}  
    \caption{Face detection results before and after being perturbed. The green, red and blue detection boxes correspond to S3FD, DSFD and RetinaFace respectively.}
    \label{fig:06}
\end{figure}

\paragraph{Disruption and Poisoning Effectiveness}
As visualized in Fig. \ref{fig:04}, the TSDF method demonstrates broadly effective interference across different attributes, images and models. The outputs generated from images protected by TSDF show significantly more distortion and artifacts compared to those from other SOTA methods, particularly on the AGGAN and StarGAN models. This demonstrates the powerful disruptive capability of the framework. Furthermore, Fig. \ref{fig:06} illustrates the success of the poisoning component. The success of TSDF's poisoning component is visually confirmed by its strong interference effect on multiple face detectors. While these detectors (S3FD, DSFD, and RetinaFace) correctly identify faces in the original images, their performance degrades substantially in protected versions by shifting their bounding boxes, reducing the number of detections or failing to detect a face.

\paragraph{Perturbation Imperceptibility}
 A crucial aspect of our framework is its ability to provide strong defense while remaining invisible to the human eye. Fig. \ref{fig:05} provides a direct visual comparison of an original image against images protected by CMUA, FOUND, and TSDF. The visual differences are nearly imperceptible, demonstrating the excellent concealment of our perturbation. This qualitative assessment is supported by the quantitative data in Table \ref{tab:table5}, which shows high PSNR and SSIM scores for the protected images across all datasets. For instance, on the CelebA dataset, TSDF achieves PSNR of 33.07 and SSIM of 0.89, demonstrating that our framework can provide a defense without compromising image imperceptibility.

\subsection{Ablation Study}

We perform ablation studies to analyze the contribution of key components of the TSDF method. Specifically, we evaluate the effectiveness of the intensity separation mechanism and the Feature Enhancement strategy.

\begin{table*}[htbp]
\centering
\caption{Impact of Perturbation Threshold on Defense Performance. This table evaluates how different threshold settings affect the trade-off between the two defensive functions. The results show the performance across various deepfake models and face detectors. Blank entries indicate that the corresponding defensive effect is non-existent.}
\setlength{\tabcolsep}{6pt} 
\small 
\begin{tabular}{cc|cccc|ccc}
\toprule
\multicolumn{2}{c|}{Configuration} & \multicolumn{4}{c|}{Deepfake Model Performance} & \multicolumn{3}{c}{Face Detection Performance} \\
\cmidrule(lr){1-2} \cmidrule(lr){3-6} \cmidrule(lr){7-9}
Threshold & Metric & HiSD & AttGAN & StarGAN & AGGAN & RetinaFace & S3FD & DSFD \\
\midrule
\multirow{5}{*}{0.00} 
    & L2mask↑ & 0.18 & 0.20 & 0.38 & 0.18 & - & - & - \\
    & SRmask↑ & 99.59\% & 99.54\% & 100.00\% & 100.00\% & - & - & - \\
    & PSNR↓ & 15.41 & 15.29 & 9.67 & 17.37 & - & - & - \\
    & SSIM↓ & 0.74 & 0.56 & 0.13 & 0.68 & - & - & - \\
    & F1-score↓ & - & - & - & - & 0.99 & 1.00 & 0.99 \\
\cmidrule(r){1-9}
\multirow{5}{*}{0.02} 
    & L2mask↑ & 0.18 & 0.21 & 0.40 & 0.19 & - & - & - \\
    & SRmask↑ & 99.45\% & 99.57\% & 100.00\% & 100.00\% & - & - & - \\
    & PSNR↓ & 15.45 & 15.29 & 9.42 & 17.04 & - & - & - \\
    & SSIM↓ & 0.74 & 0.55 & 0.13 & 0.67 & - & - & - \\
    & F1-score↓ & - & - & - & - & 0.74 & 0.99 & 0.99 \\
\cmidrule(r){1-9}
\multirow{5}{*}{0.03}
    & L2mask↑ & 0.17 & 0.20 & 0.39 & 0.18 & - & - & - \\
    & SRmask↑ & 99.22\% & 99.56\% & 100.00\% & 100.00\% & - & - & - \\
    & PSNR↓ & 15.83 & 15.52 & 9.47 & 17.47 & - & - & - \\
    & SSIM↓ & 0.75 & 0.56 & 0.13 & 0.68 & - & - & - \\
    & F1-score↓ & - & - & - & - & 0.75 & 0.84 & 0.99 \\
\cmidrule(r){1-9}
\multirow{5}{*}{0.04}
    & L2mask↑ & 0.12 & 0.17 & 0.37 & 0.18 & - & - & - \\
    & SRmask↑ & 99.15\% & 98.25\% & 100.00\% & 100.00\% & - & - & - \\
    & PSNR↓ & 16.14 & 16.76 & 9.73 & 17.86 & - & - & - \\
        & SSIM↓ & 0.76 & 0.62 & 0.17 & 0.70 & - & - & - \\
    & F1-score↓ & - & - & - & - & 0.77 & 0.23 & 0.29 \\
\cmidrule(r){1-9}
\multirow{5}{*}{0.05}
    & L2mask↑ & 0.01 & 0.01 & 0.14 & 0.13 & - & - & - \\
    & SRmask↑ & 0.00\% & 0.00\% & 98.10\% & 99.22\% & - & - & - \\
    & PSNR↓ & 30.69 & 31.50 & 14.80 & 22.10 & - & - & - \\
    & SSIM↓ & 0.94 & 0.94 & 0.46 & 0.83 & - & - & - \\
    & F1-score↓ & - & - & - & - & 0.97 & 0.03 & 0.05 \\
\bottomrule
\end{tabular}
\label{tab:table6}
\end{table*}

\begin{table}[htbp]
\centering
\small  
\setlength{\tabcolsep}{1pt}  
\caption{Compared with different feature distance metrics strategies.}
\begin{tabular}{@{}llcccc@{}}
\toprule
Method & Model & L2mask$\uparrow$ & SRmask$\uparrow$ & PSNR$\downarrow$ & SSIM$\downarrow$ \\
\midrule
\multirow{4}{*}{None} 
 & HiSD    & 0.06 & 73.12\% & 20.12 & 0.82 \\
 & AttGAN  & 0.16 & 98.42\% & 16.98 & 0.63 \\
 & StarGAN & 0.20 & 100.00\% & 12.67 & 0.25 \\
 & AGGAN   & 0.13 & 99.90\% & 21.98 & 0.79 \\
\midrule
\multirow{4}{*}{KL Divergence} 
 & HiSD    & 0.11 & 94.14\% & 19.81 & 0.82 \\
 & AttGAN  & 0.13 & 95.40\% & 18.59 & 0.70 \\
 & StarGAN & 0.22 & 100.00\% & 12.34 & 0.19 \\
 & AGGAN   & 0.19 & 99.98\% & \textbf{16.54} & 0.71 \\
\midrule
\multirow{4}{*}{Wasserstein Distance} 
 & HiSD    & 0.10 & 93.88\% & 19.81 & 0.83 \\
 & AttGAN  & 0.13 & 95.20\% & 18.68 & 0.71 \\
 & StarGAN & 0.22 & 100.00\% & 12.39 & 0.20 \\
 & AGGAN   & \textbf{0.19} & 100.00\% & 16.55 & 0.70 \\
\midrule
\multirow{4}{*}{Feature Enhancement} 
 & HiSD    & \textbf{0.18} & \textbf{99.59\%} & \textbf{15.41} & \textbf{0.74} \\
 & AttGAN  & \textbf{0.20} & \textbf{99.54\%} & \textbf{15.29} & \textbf{0.56} \\
 & StarGAN & \textbf{0.38} & \textbf{100.00\%} & \textbf{9.67} & \textbf{0.13} \\
 & AGGAN   & 0.18 & \textbf{100.00\%} & 17.37 & \textbf{0.68} \\
\bottomrule
\end{tabular}
\label{tab:table7}
\end{table}

\paragraph{Impact of the Intensity Separation Threshold}

The ability of TSDF to combine two defensive functions relies on the intensity separation mechanism, which is controlled by a threshold $\tau$. To find the optimal balance, we tested various threshold values, with the results presented in Table \ref{tab:table6}. The data show a clear functional conflict between the interruption and poisoning effects. At a low threshold such as 0.02, the framework dedicates more of the perturbation's budget to interruption. This results in strong performance against deepfake models, achieving L2mask of 0.40 on StarGAN, but leads to a weaker poisoning effect on face detectors with F1-score of 0.99 on S3FD. Conversely, a high threshold like 0.05 allocates more space to poisoning, which significantly impairs face detectors by reducing the F1-score on S3FD to 0.03, but this is at the expense of weakening the interruption effect. Our experiments identify a threshold of 0.04 as the optimal balance, providing balanced performance in both interruption and poisoning across all models and detectors. This validates the effectiveness of the separation strategy.

To validate that the framework's poisoning efficacy originates from our dedicated component, we evaluate an interruption-only variant of TSDF ($\tau$ = 0.00) in which the poisoning module is completely deactivated. The results are clear. The interruption-only perturbation of TSDF has almost no impact on the face detectors' F1-score, which remained nearly the same as the undefended baseline. In contrast, the full TSDF method significantly reduces the F1-score. This proves that the strong poisoning capability comes from the poisoning component and is not the effect of interruption.

\paragraph{Effectiveness of the Feature Enhancement Strategy}

To isolate and evaluate the contribution of our Feature Enhancement strategy specifically to the interruption component, we conduct a dedicated ablation study. In this experiment, the poisoning module is deactivated across all test configurations to ensure that its effects do not interfere with the comparison of interruption performance. We then compare it with several alternative feature distance metrics, using the Kullback-Leibler (KL) Divergence and the Wasserstein Distance \cite{b29}. As shown in Table \ref{tab:table7}, the Feature Enhancement method consistently yields the best interruption performance. As shown in the case of StarGAN, the proposed strategy achieves an L2mask score of 0.38. This result surpasses that of the KL Divergence method, where the corresponding L2mask value is only 0.22. The baseline without enhancement performs even lower. This confirms that the Feature Enhancement strategy is more effective in creating powerful and disruptive perturbations. These results demonstrate the superior efficacy of our Feature Enhancement strategy, which produces a significantly more disruptive perturbation than the baselines. The high efficacy of this primary interruption is a key requirement for our two-stage paradigm, as it establishes the foundational defense that the poisoning component is designed to make persistent.

\section*{Conclusion}

To address the critical lack of persistence in active defenses, this paper introduces the Two-Stage Defense Framework, a novel approach that synergistically combines interruption and data poisoning. This fusion is achieved through the proposed intensity separation mechanism by integrating both functions into a single perturbation.
The method provides a powerful and real-time interruption by distorting forged results at the inference stage. Meanwhile, the poisoning component ensures long-term persistence by interfering with the face detection step in an attacker’s retraining process. This directly addresses the fundamental vulnerability of conventional defenses, which inevitably fail against such adaptive attacks. Experimental results show that our framework effectively distorts the output of the mainstream deepfake model, while also demonstrating a strong capability to contaminate the face detection results. Ultimately, this work establishes a new paradigm for persistent active defense, moving beyond passively surviving retraining to actively destroying an attacker's ability to adapt.

\end{document}